%% file: main_arxiv.tex
\PassOptionsToPackage{table}{xcolor}

\documentclass[sigconf]{acmart}

\AtBeginDocument{%
  }

\usepackage{hyperref}
\usepackage{url}
\usepackage{wrapfig}
\hypersetup{
    colorlinks,
    linkcolor={red!50!black},
    citecolor={blue!50!black},
    urlcolor={blue!80!black}
}

\usepackage[T1]{fontenc}
\usepackage[utf8]{inputenc}
\usepackage{microtype}

\usepackage{amsmath, graphicx}
\usepackage{enumitem}
\usepackage[table]{xcolor}

\usepackage{pifont} 

\usepackage{booktabs}       
\usepackage{amsfonts}       
\usepackage{nicefrac}       
\usepackage{microtype}      
\usepackage{multirow}
\usepackage{soul}
\usepackage{balance}
\usepackage{tcolorbox}

\usepackage[linesnumbered,ruled,vlined]{algorithm2e}
\SetKwInput{KwInput}{Input}                
\SetKwInput{KwOutput}{Output}              

\newcommand{\by}{\boldsymbol{y}}
\newcommand{\bs}{\boldsymbol{s}}
\newcommand{\bx}{\boldsymbol{x}}
\newcommand{\monospace}[1]{{\fontfamily{qcr}\selectfont #1}}

\usepackage{mathtools}
\definecolor{rowblue}{RGB}{230, 245, 255}

\copyrightyear{2025}
\acmYear{2025}
\setcopyright{cc}
\setcctype{by}
\acmConference[KDD '25]{Proceedings of the 31st ACM SIGKDD Conference on Knowledge Discovery and Data Mining V.2}{August 3--7, 2025}{Toronto, ON, Canada}
\acmBooktitle{Proceedings of the 31st ACM SIGKDD Conference on Knowledge Discovery and Data Mining V.2 (KDD '25), August 3--7, 2025, Toronto, ON, Canada}
\acmDOI{10.1145/3711896.3736871}
\acmISBN{979-8-4007-1454-2/2025/08}

\begin{document}

\title{Calibrating Pre-trained Language Classifiers on LLM-generated Noisy Labels via Iterative Refinement}

\author{Liqin Ye}
\email{liqiny@gatech.edu}
\orcid{1234-5678-9012}
\affiliation{%
  \department{School of Computational Science \& Engineering}
  \institution{Georgia Institute of Technology}
  \city{Atlanta}
  \state{GA}
  \country{USA}
}

\author{Agam Shah}
\email{ashah482@gatech.edu}
\affiliation{%
  \department{School of Computational Science \& Engineering}
  \institution{Georgia Institute of Technology}
  \city{Atlanta}
  \state{GA}
  \country{USA}
}

\author{Chao Zhang}
\email{chaozhang@gatech.edu}
\affiliation{%
  \department{School of Computational Science \& Engineering}
  \institution{Georgia Institute of Technology}
  \city{Atlanta}
  \state{GA}
  \country{USA}
}

\author{Sudheer Chava}
\email{sudheer.chava@scheller.gatech.edu}
\affiliation{%
  \department{Scheller College of Business}
  \institution{Georgia Institute of Technology}
  \city{Atlanta}
  \state{GA}
  \country{USA}
}


\input{sections/00_abs}

\ccsdesc[500]{Computing methodologies~Natural language processing}

\keywords{Large Language Model, Noisy Labels, Diffusion Model}

\maketitle

\input{sections/01_intro}
\input{sections/02_background}
\input{sections/03_method}

\input{sections/04_exp}
\input{sections/05_relate}
\input{sections/06_conclusion}

\begin{acks}
    This work was supported in part by NSF IIS-2008334, IIS-2106961, IIS-2403240, and CAREER IIS-2144338. We sincerely thank Arnav Hiray, Michael Galarnyk and the anonymous reviewers for their valuable comments and constructive suggestions, which significantly improved the quality of this paper. We gratefully acknowledge the Partnership for an Advanced Computing Environment (PACE) at the Georgia Institute of Technology for providing computing resources that enabled this research.
\end{acks}

\bibliographystyle{ACM-Reference-Format}
\bibliography{ref}
\vspace{-0.7em}
\input{sections/appendix}

\end{document}

%% file: sections/00_abs.tex
\begin{abstract}
  The traditional process of creating labeled datasets is labor-intensive and expensive. Recent breakthroughs in open-source large language models (LLMs) have opened up a new avenue in generating labeled datasets automatically for various natural language processing (NLP) tasks, providing an alternative to such an expensive annotation process. However, the reliability of such auto-generated labels remains a significant concern due to inherent inaccuracies. When learning from noisy labels, the model's generalization is likely to be harmed as it is prone to overfit to those label noises. While previous studies in learning from noisy labels mainly focus on synthetic noise and real-world noise, LLM-generated label noise receives less attention. In this paper, we propose \textbf{SiDyP}: \textbf{Si}mplex Label Diffusion with \textbf{Dy}namic \textbf{P}rior to calibrate the classifier's prediction, thus enhancing its robustness towards LLM-generated noisy labels. SiDyP retrieves potential true label candidates by neighborhood label distribution in text embedding space and iteratively refines noisy candidates using a simplex diffusion model. Our framework can increase the performance of the BERT classifier fine-tuned on both zero-shot and few-shot LLM-generated noisy label datasets by an average of 7.21\% and 7.30\% respectively. We demonstrate the effectiveness of SiDyP by conducting extensive benchmarking for different LLMs over a variety of NLP tasks. Our code is available on GitHub\footnote{https://github.com/gtfintechlab/SiDyP}.  
\end{abstract}

%% file: sections/01_intro.tex
\section{Introduction}
In the era of advanced LLMs, the capabilities for automatic data annotation have seen remarkable improvements \citep{wang2024surveydatasynthesisaugmentation, Gilardi_2023, wang2023tsciqteachingmultimodalchainofthought, zhu2023chatgptreproducehumangeneratedlabels, tan2024largelanguagemodelsdata, yu2023explanation, brown2020languagemodelsfewshotlearners}. LLMs, leveraging their extensive training on diverse textual data, can annotate data more efficiently and cost-effectively compared to traditional scalable labeling methods, such as crowdsourcing \citep{sakaguchi2019winograndeadversarialwinogradschema, williams2018broadcoveragechallengecorpussentence}, labeling rules \citep{zhang2021wrench} and web annotations \citep{goh2018usingrulebasedlabelsweak}. Due to the looming data exhaustion crisis, LLM synthesis datasets have become increasingly prevalent in contemporary research and applications \citep{wang2024surveydatasynthesisaugmentation, li2023syntheticdatagenerationlarge, Hastings_2024}. Despite extensive efforts to enhance the accuracy and reliability of LLM-annotated labels \citep{yu2023explanation, yu-bach-2023-alfred, wang2023noiserobustfinetuningpretrainedlanguage, oliveira2024combining, li2024openworldmultilabeltextclassification, burns2023weaktostronggeneralizationelicitingstrong}, label noises (incorrect labels) remain inevitable \citep{snorkelUsingFewshot, qin2023chatgptgeneralpurposenaturallanguage, shah2023zeroheroyetbenchmarking}. Deep Neural Networks (DNNs) are susceptible to those noise as they tend to inadvertently fit inaccuracies \citep{arpit2017closerlookmemorizationdeep, cheng2021learninginstancedependentlabelnoise, zhang2021learningnoisetransitionmatrix}. Hence, it necessitates a robust mechanism to mitigate the harmful impact of these label noises.  

\begin{figure*}[h]
    \centering
    \includegraphics[width=0.9\textwidth]{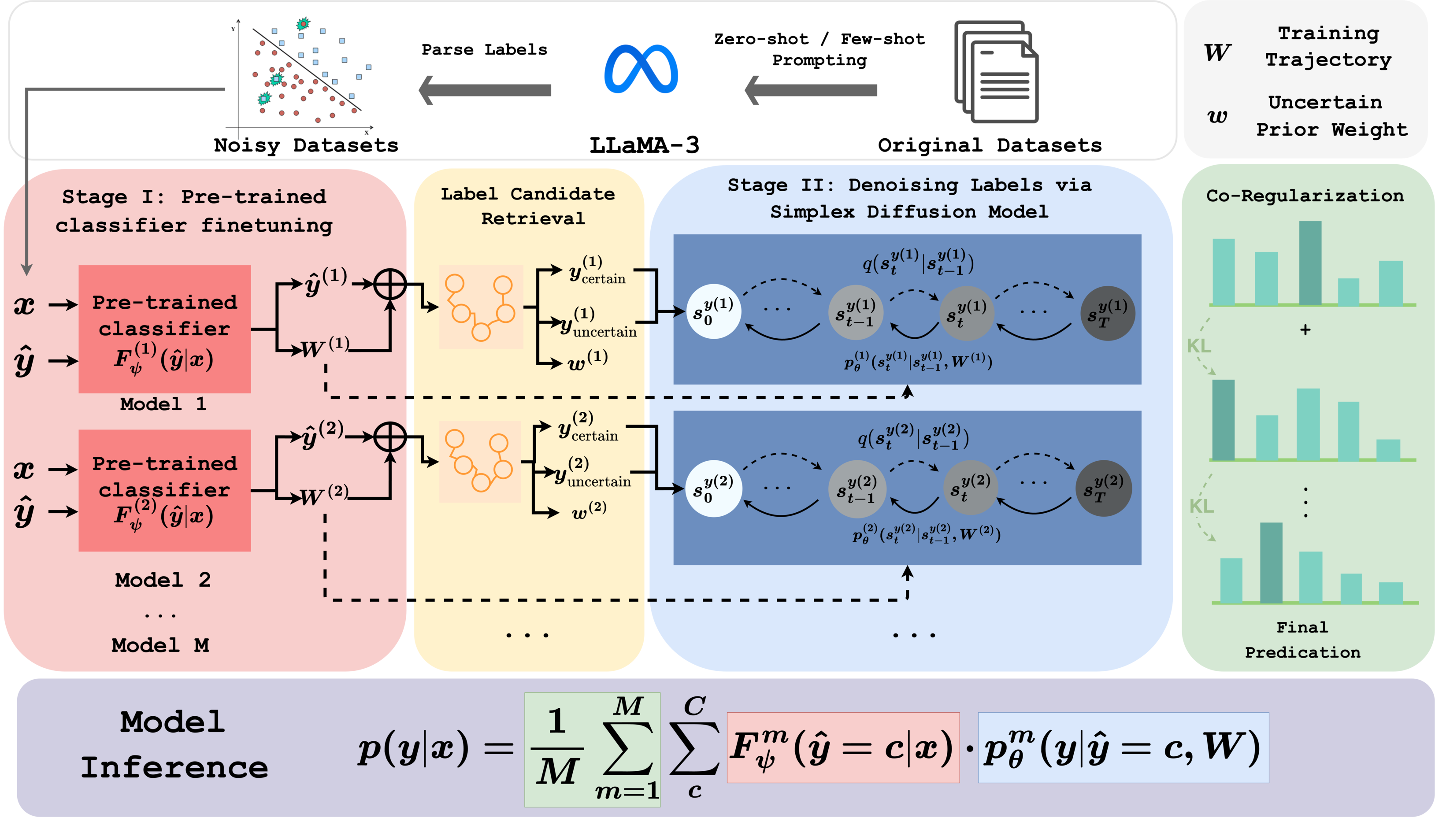}
    \caption{The SiDyP framework, containing (1) pre-trained classifier fine-tuning; (2) dynamic label candidates retrieval and distillation; (3) denoising label using simplex diffusion; (4) co-regularization between multiple model branches; (5) inference process to predict refined labels from noisy labels. }
    \label{fig:sidyp_pipeline}
\end{figure*}

Learning from noisy labels has been extensively studied. A variety of techniques have been proposed to mitigate the adverse effects of label noise on DNNs (data cleaning, regularization, noise transition estimation, etc.) \citep{arazo2019unsupervisedlabelnoisemodeling, zhuang2023dygen, han2018coteaching, bae2022noisy, nguyen2019selflearningfilternoisy, wei2020combating, zhang2018generalizedcrossentropyloss, yu2019doesdisagreementhelpgeneralization, yao2022instancedependentlabelnoiselearningstructural, xia2020partdependentlabelnoiseinstancedependent}. However, most work concentrates on either synthetic noise, whose class‑dependent or homogeneous corruption does not capture real annotation errors \citep{wei2022learningnoisylabelsrevisited}, or on real‑world noisy datasets, which are expensive to construct—requiring expert‑crafted labeling functions \citep{ratner2017snorkel} or large‑scale crowdsourcing. The majority of the studies focus on either synthetic noise or real-world noise \citep{food, cloth}. Given the extensive research on improving LLM annotation ability and its promising efficacy in substituting traditional tedious labeling processes, LLM-generated noise remains largely unexplored. To bridge this gap, we propose an innovative denoising approach SiDyP that strengthens classifiers' resilience to LLM-generated noisy labels. We benchmark SiDyP and previous state-of-the-art learning from noisy label methods on different LLMs for various NLP tasks.

SiDyP aims to calibrate noisy labels using transition matrix-based methods \citep{patrini2017making, yao2021dual, zhang2021learning, xia2020partdependent, berthon2021confidence}. Our denoising method consists of two stages: finetuning pre-trained language classifiers (PLCs) and denoising via diffusion models. Finetuning a PLC on a noisy dataset yields training dynamics, the trajectories in embedding space during training \citep{zhuang2023dygen}. Observing that LLM-generated label noises are more intricate and context-dependent (See section \ref{definition}), we collect a list of potential true label candidates instead of a fixed corresponding true label by referring to the neighbor's label distribution in embedding space. We design a simplex diffusion \citep{mahabadi2024tess} label model to reconstruct true labels from noisy labels conditioned on training dynamics. The potential true label candidates are refined progressively throughout the training of the diffusion model based on its prediction. The overall framework is presented in Figure \ref{fig:sidyp_pipeline}.

The main contributions of our work are summarized as follows:
\begin{itemize}
    \item We evaluate previous state-of-the-art baselines in the problem of learning from noisy labels under a novel type of noise: LLM-generated label noise. To the best of our knowledge, this is the first study aimed at enhancing learning under LLM-generated label noise.
    \item We propose SiDyP, a framework correcting the classifier's prediction by using a simplex denoising label diffusion model to progressively refine the noisy labels. To address the challenges posed by LLM-generated noise, a more context-dependent noise,  we design a label-candidate retrieval algorithm.
    \item We conduct extensive experiments of our frameworks compared to 5 state-of-the-art baselines across 4 NLP tasks, 5 LLMs, and 3 different types of noises. Our approach outperforms all baselines in all experiments. The effectiveness of each component is also verified and analyzed.
\end{itemize}

%% file: sections/02_background.tex
\section{Background and Motivation}
\label{definition}

\begin{figure*}[h]
    \includegraphics[width=0.94\textwidth]{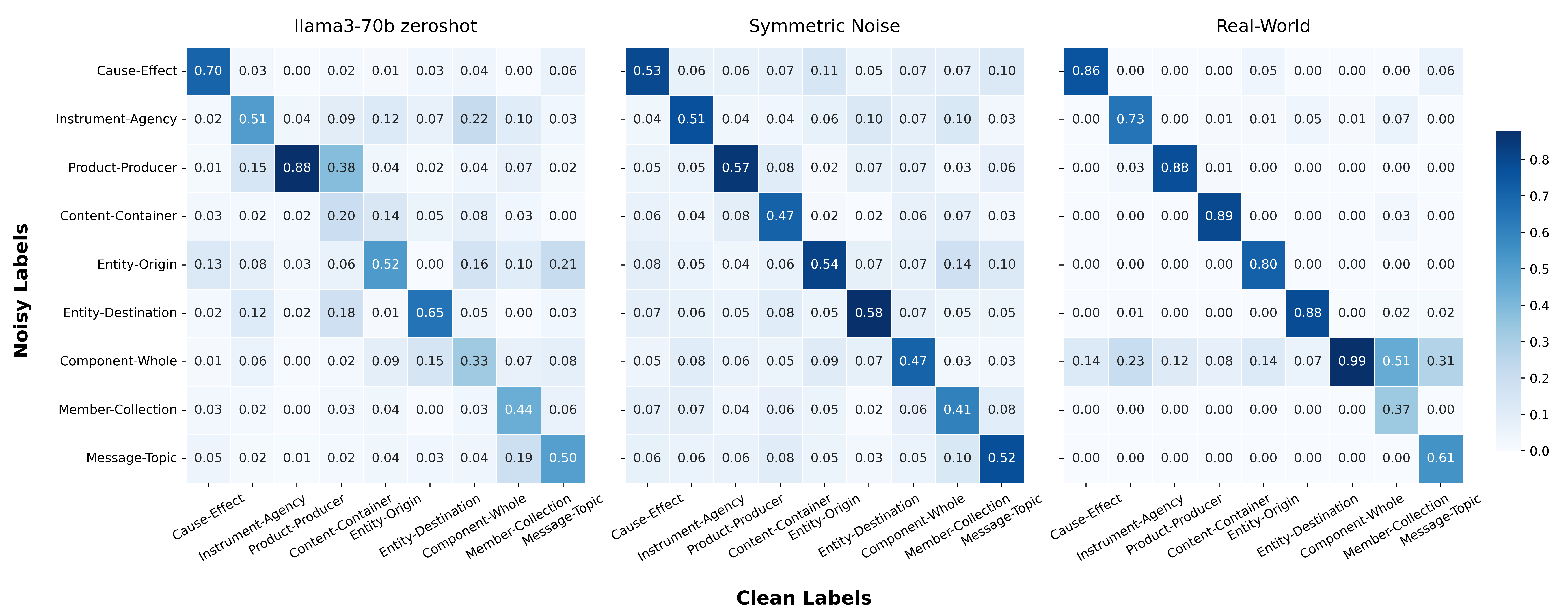}
    \caption{Confusion Matrix of LLM-generated label noise, synthetic noise, and real-world noise on SemEval dataset. We prompt Llama-3-70b in zero-shot fashion to gather LLM-generated labels. We inject symmetric noise to obtain synthetic noise. Real-world labels are collected by 164 labeling functions written by subject matter experts \citep{ratner2017snorkel}.}
    \label{fig:noise_character}
\end{figure*}

\paragraph{Problem Definition} Let $\mathcal{X} \in \mathbb{R}^d$ and $\mathcal{Y}=\{0,1,...,c-1\}$ be the $d$-dimension input and the target label in a classification task with $c$ classes. Following the joint probability distribution $P$ over $\mathcal{X} \times \mathcal{Y}$, the i.i.d. samples form a gold classification dataset, $\mathcal{D}=\{x_i,y_i\}^{N}_{i=1}$. Our assumption of learning from noisy labels indicates that the only accessible dataset is $\mathcal{\Tilde{D}_{\text{train}}}=\{x_i, \Tilde{y}_i\}^{N}_{i=1}$, sampled from $\Tilde{P}$ over $\mathcal{X} \times \mathcal{\Tilde{Y}}$ where $\mathcal{\Tilde{Y}}$ are potential noisy targets. For a traditional classification problem, the training objective of a classifier $f_{\theta}$ is to minimize the true risk $R_L(f_{\theta}):=\mathbb{E}_P[L(f_{\theta}(x), y)]$ where $L(\cdot)$ is the loss function. However, in the realm of learning from noisy labels, the only accessible risk function is the noisy empirical risk $\Tilde{R}^{\text{emp}}_{L}(f_{\theta}):=\mathbb{E}_P[L(f_{\theta}(x), \Tilde{y})]$ due to the absence of true labels $y$. Therefore, our goal is to find a function minimizing the true risk $R_L(f_{\theta})$ during learning with noisy empirical risk $\Tilde{R}^{\text{emp}}_L(f_{\theta})$.
With the only observable target labels being noisy, we manage to train a model that generates the probability distribution of true label $y$ given arbitrary input $x$, $p(y|x)$. Taking advantage of noisy labels in our training dataset, we can decompose our objective further as:
$$p(y|x)=\sum\limits_{\Tilde{y}}p(\Tilde{y}|x)p(y|\Tilde{y},x)$$

In this revised objective, the prior $p(\Tilde{y}|x)$ can be directly estimated by finetuning a PLC $\boldsymbol{F_{\psi}}$ on the accessible noisy dataset. We can approximate the posterior $p(y|\Tilde{y}, x)$, expressing the probability distribution of true label $y$ given noisy label $\Tilde{y}$ and input $x$, by a generative model.

\paragraph{Motivation} The emergence of LLMs makes automatic annotations feasible, easing the burdens of tedious manual annotations. Its performance in text annotations exceeds crowd-workers by an average of 25\% while at a cost of 30 times cheaper than MTurk \citep{Gilardi_2023}. However, their generated labels are not error-free \citep{snorkelUsingFewshot, qin2023chatgptgeneralpurposenaturallanguage}. Training DNNs on these noisy labels leads to deficient performance. Previous studies in the realm of learning from noisy labels focus heavily on benchmarking synthetic noise and real-world noise \citep{nguyen2019selflearningfilternoisy, yu2019doesdisagreementhelpgeneralization,xia2020partdependentlabelnoiseinstancedependent}. LLM-generated label noise has received insufficient attention. To make DNNs robust to LLM-generated label noises, we need to first understand the differences between LLM-generated label noises and other widely benchmarked noises (synthetic and real-world). Figure \ref{fig:noise_character} and Figure \ref{fig:noise_character_full} presents the transition matrix of SemEval \citep{hendrickx2019semeval}, a semantic-relationship dataset, under these three types of noise: LLM, synthetic, and real-world. We explore three popular synthetic noises: Symmetric Noise, Asymmetric Noise, and Instance-Dependent Noise (See details in Section \ref{syn_real_noise}). Except for real-world noise which has a lower noise ratio (16\%), both LLM-generated noises and synthetic noise's ratio are around 50\%. We observe the following:
\begin{itemize}[leftmargin=*]
    \item Despite that LLM-generated labels have a similar noise ratio with the synthetic noise, its correct label percentage (the diagonal) is more diverse. In contrast, synthetic noises have an individual class noise ratio similar to the total noise ratio (50\%).
    \item In synthetic noises, incorrect labels often show clear patterns: ASN label is consistently off by one class (See Figure \ref{fig:noise_character_full} in Appendix \ref{ap:noise_dist}), and SN distributed relatively equally. The label noise introduced by IDN changes significantly depending on the seed used (See Figure \ref{fig:noise_character_full} in Appendix \ref{ap:noise_dist}). Such a sensitivity to the initial random state impacts model robustness.
    \item While the distribution of synthetic noise indicates that this type of mislabeling often lacks contextual correlation, LLM-generated label noise reflects underlying relationships between classes (as evidenced by the similarity among the three LLMs. See Figure \ref{fig:noise_character_full} in Appendix \ref{ap:noise_dist}), making it more aligned with real-world noise.
\end{itemize}
These observations trigger a more challenging estimation of the posterior as the relation between $\tilde{y}$ and $y$ becomes less predictable and more context-dependent. To tackle this, we begin by focusing on these two key aspects: 
\begin{enumerate}
    \item How can a promising and reliable true label be derived from the noisy dataset?
    \item How can we estimate such a probabilistic relation between true labels, mislabeled labels, and input features accurately?
\end{enumerate}

In the following sections, we introduce our true label candidates dynamic distillation (Section \ref{dy_distill}) and simplex denoising label diffusion model (Section \ref{simplex_diff}) to address these two concerns respectively. We also adopt training dynamics during PLC fine-tuning and co-regularization mechanism (Appendix \ref{td_cr}) to make SiDyP tolerant to noises.

%% file: sections/03_method.tex
\section{True Label Candidates Distillation}
\label{dy_distill}
Extracting true labels from a noisy dataset is crucial, as it directly impacts the quality of the subsequent generative posterior approximation. Our true label derivation is based on the assumption that textual embeddings are robust enough to discriminate between clean and corrupted data samples \citep{ortego2021multi}. Texts belonging to the same class typically exhibit similar semantics, making them more likely to cluster together in the embedding space. Therefore, the neighboring labels reveal information about the true labels. Different from prior works \citep{zhuang2023dygen, bae2022noisy}, we retrieve a list of true label candidates for each individual data sample (Algorithm \ref{alg:retrieval}). These true label candidates are distilled according to our diffusion model's feedback during training (Algorithm \ref{alg:distillation}).

\begin{algorithm}[!ht]
\DontPrintSemicolon
  
  \KwInput{$\mathcal{D}^{\text{noisy}}_{\text{train}}$: $\{\bf{x_i}, \bf{\tilde{y}_i}\}^{n}_{i}$, $\mathcal{M}_{\text{train}}$, $\mathcal{C}_{\text{knn}}$, $K,\lambda,\gamma$}
  \KwOutput{$\mathcal{D}^{\text{certain}}_{\text{train}}$: $\{\bf{x_i}, \bf{y}_i\}^{m}_{i}$, $\mathcal{D}^{\text{uncertain}}_{\text{train}}$: $\{\bf{x_i}, (\bf{y}^0_i, \bf{y}^1_i, \dots)\}^{n-m}_{i}$, $\mathcal{W}^{\text{uncertain}}_{\text{train}}$: $\{(\bf{w}^0_i, \bf{w}^1_i, \dots)\}^{n-m}_{i}$}
  Split $\mathcal{D}^{\text{noisy}}_{\text{train}}$ into \{$\mathcal{\bar{D}}^{\text{clean}}_{\text{train}}$, $\mathcal{\bar{D}}^{\text{noisy}}_{\text{train}}$\} according to noisy marker $\mathcal{M}_{\text{train}}$\\
  Fit $\mathcal{\bar{D}}^{\text{clean}}_{\text{train}}$ into KNN classifier $\mathcal{C}_{\text{knn}}$ \\
  Predict $\mathcal{P}_{\text{train}}:\{(\bf{p}^0_i, \bf{p}^1_i,\dots)\}^{n}_{i}$ of entire dataset $\mathcal{D}^{\text{noisy}}_{\text{train}}$ using $\mathcal{C}_{\text{knn}}$ based on $K$ neighbors\\
  Initialize $\mathcal{D}^{\text{certain}}_{\text{train}}=\{\}, \mathcal{D}^{\text{uncertain}}_{\text{train}}=\{\}$ and $\mathcal{W}^{\text{uncertain}}_{\text{train}}$=\{\} \\
  \For{$i=0$ to $n$}
    {
        $\bf{p}^{max}_i$ = max$\{(\bf{p}^0_i, \bf{p}^1_i,\dots)\}$ \\
        \If{$\bf{p}^{max}_i \geq \lambda$}
            {   
                Insert $(\bf{x_i},\bf{y}^{max}_i)$ into $\mathcal{D}^{\text{certain}}_{\text{train}}$
            }
        \Else
            {
                $\bf{p}^{max1}_i,\bf{p}^{max2}_i$ = top2$\{(\bf{p}^0_i, \bf{p}^1_i,\dots)\}$ \\
                \If {$\bf{p}^{max1}_i+\bf{p}^{max2}_i \geq \gamma$}
                    {
                        Insert $(\bf{x_i},\{\bf{y}^{max1}_i,\bf{y}^{max2}_i\})$ into $\mathcal{D}^{\text{uncertain}}_{\text{train}}$ \\
                        $\bf{p}^{max1}_i,\bf{p}^{max2}_i$ = normalize($\bf{p}^{max1}_i,\bf{p}^{max2}_i$) \\
                        Insert $(\bf{p}^{max1}_i,\bf{p}^{max2}_i)$ into $\mathcal{W}^{\text{uncertain}}_{\text{train}}$
                    }
                \Else
                    {
                        Insert $(\bf{x_i},\{\bf{y}^0_i,\bf{y}^1_i, \dots\})$ into $\mathcal{D}^{\text{uncertain}}_{\text{train}}$ \\
                        Insert $(\bf{p}^0_i,\bf{p}^1_i, \dots)$ into $\mathcal{W}^{\text{uncertain}}_{\text{train}}$
                    }
            }
    }
\caption{Potential True Label Candidates Retrieval}
\label{alg:retrieval}
\end{algorithm}

\subsection{Label Candidate Retrieval (Algorithm \ref{alg:retrieval})}
One of our main purposes is to discriminate noisy samples in the dataset and obtain clean label information. During the PLC fine-tuning in Stage I, there exist training dynamics in embedding space. Noisy samples tend to exhibit larger mean and standard deviation of Euclidean distances towards their assigned labels (incorrect) compared to clean samples \citep{zhuang2023dygen}. Hence, we can split the original dataset into $D^{\text{noisy}}_{\text{train}}$ and $D^{\text{clean}}_{\text{train}}$ by cutting off the top $\sigma$ percent of training trajectories, where $\sigma$ is the estimated error rate. We apply K-Nearest Neighbor (KNN) algorithm on $D^{\text{noisy}}_{\text{train}}$ with $D^{\text{clean}}_{\text{train}}$ as a reference. Instead of assigning a single deterministic label, a list of label candidates and their corresponding weights (probability) are generated by the KNN classifier. We manage to alleviate the uncertainty injected into the training of the diffusion model in Stage II by two filters:
\begin{itemize}[leftmargin=*]
    \item We preserve the candidate if its associated probability is greater than a threshold $\lambda$. These data instances are regarded as deterministic instances since their potential true label is single and certain. The remaining data instances are regarded as uncertain and linked with a list of candidates.
    \item For uncertain data instances, we extract the two candidates with the highest probabilities. If their summation is greater than a specified threshold $\gamma$, we then eliminate other candidates and only preserve these two dominant candidates.
\end{itemize}

\subsection{Candidate Distillation (Algorithm \ref{alg:distillation})}
As we collect a list of label candidates, it inevitably introduces uncertainty for generative modeling, leading to degenerate performance. To mitigate this, we first train our generative model only on the deterministic dataset for $\alpha$ warm-up epochs. We use this model to evaluate our uncertain dataset over a specified iteration $\beta$. During each evaluation, if the model's predicted label lies in the candidate list, the matched label candidate will increase accordingly. The weight list will then be normalized as well to maintain a sum of 1. After the candidate weight update and model evaluation for uncertain data samples, we sample a specific label candidate from the candidate list multinomially based on the candidate weights. We treat such a sample label as the true label in this training epoch. The generative model is then trained on both deterministic pairs and uncertain pairs. Subsequently, the loss of the generative model for an uncertain sample is weighted by the sampled candidate's weight.

\begin{algorithm}[!ht]
\DontPrintSemicolon
  
  \KwInput{$\mathcal{G}_{\text{model}}$, $\mathcal{D}^{\text{certain}}_{\text{train}}$: $\{\bf{x_i}, \bf{y}_i\}^{m}_{i}$, $\mathcal{D}^{\text{uncertain}}_{\text{train}}$: $\{\bf{x_i}, (\bf{y}^0_i, \bf{y}^1_i, \dots)\}^{n-m}_{i}$, $\mathcal{W}^{\text{uncertain}}_{\text{train}}$: $\{(\bf{w}^0_i, \bf{w}^1_i, \dots)\}^{n-m}_{i}$, $\alpha$, $E$, $\beta$}
  \KwOutput{$\mathcal{G}_{\text{model}}$}
  \For{$e=0$ to $E$}
    {
        \If{$e \leq \alpha$}
            {
                 $\{\bf{\bar{y}_i}\}^{m}_{i}$ = $\mathcal{G}_{\text{model}}[\{\bf{x_i}\}^{m}_{i}]$ for $\mathcal{D}^{\text{certain}}_{\text{train}}$\\
                loss = $\mathcal{F}_{\text{loss}}[\{\bf{\bar{y}_i}\}^{m}_{i}, \{\bf{y}_i\}^{m}_{i}]$ \\
                Optimize $\mathcal{G}_{\text{model}}$
            }
        \Else
            {
                \For{$i=0$ to $\beta$}
                    {
                        $\{\bf{\bar{y}_i}\}^{n-m}_{i}$ = $\mathcal{G}_{\text{model}}[\{\bf{x_i}\}^{n-m}_{i}]$ for $\mathcal{D}^{\text{uncertain}}_{\text{train}}$ \\
                        \If{$\{\bf{\bar{y}_i}\}^{n-m}_{i}$ in $(\bf{y}^0_i, \bf{y}^1_i, \dots)$}
                            {
                                Increase corresponding $\bf{w^*_i}$ by $\frac{1-\bf{w^*_i}}{\beta}$ \\
                                $(\bf{w}^0_i, \bf{w}^1_i, \dots)$ = normalize[$(\bf{w}^0_i, \bf{w}^1_i, \dots)$]
                            }
                    }
                $\{\bf{y_i}\}^{n-m}_{i}$= sample $(\bf{y}^0_i, \bf{y}^1_i, \dots)$ multinomially according to $\mathcal{W}^{\text{uncertain}}_{\text{train}}$ \\
                $\{\bf{\bar{y}_i}\}^{n-m}_{i}$ = $\mathcal{G}_{\text{model}}[\{\bf{x_i}\}^{n-m}_{i}]$ for $\mathcal{D}^{\text{uncertain}}_{\text{train}}$ \\
                $\{\bf{\bar{y}_i}\}^{m}_{i}$ = $\mathcal{G}_{\text{model}}[\{\bf{x_i}\}^{m}_{i}]$ for $\mathcal{D}^{\text{certain}}_{\text{train}}$ \\
                certain\_loss = $\mathcal{F}_{\text{loss}}[\{\bf{\bar{y}_i}\}^{m}_{i}, \{\bf{y}_i\}^{m}_{i}]$ \\
                uncertain\_loss = $\{\bf{\bar{w}_i}\}^{n-m}_{i} \times \mathcal{F}_{\text{loss}}[\{\bf{\bar{y}_i}\}^{n-m}_{i}, \{\bf{y}_i\}^{n-m}_{i}]$ \\
                loss = certain\_loss + uncertain\_loss \\
                Optimize $\mathcal{G}_{\text{model}}$
            }
    }
    
\caption{Distill True Label from Candidates}
\label{alg:distillation}
\end{algorithm}

\section{Simplex Denoising Label Diffusion Model}
\label{simplex_diff}
In terms of posterior approximation via generative models, we tackle it from the perspective of denoising diffusion models, which are designed for reconstructing high-fidelity data from pure noise iteratively. We view true label inference as a progressive denoising process from the noisy label based on input feature $x$. In this paper, we apply the simplex diffusion model \citep{mahabadi2024tess}, one of the continuous diffusion models, to approximate the true label posterior probability from noisy labels. Simplex diffusion models diffuse in simplex probability space, which aligns with our attempt to estimate the posterior distribution.

\paragraph{Label Simplex Representation} True label $y$ will be represented in one-hot encoded format $y \in \{0,1\}^{C}$. For a specific class $c$, $y_c = 1$ and $y_i=0$ where $i \neq c$. Given the discrete nature of one-hot data representation, we need to first map such categorical data to continuous space to fit our continuous simplex diffusion model. We map the one-hot label representation $y \in \{0,1\}^C$ to $k$-logit simplex to generate $s^y \in \{\pm k\}^{|C|}$, whose $i$-th component satisfies 
\begin{equation}
    s^{c}_{(i)}=
        \begin{cases} 
            k, &\text{if } i = c, \\
            -k &\text{otherwise.}
        \label{eq1}
        \end{cases}
\end{equation}

where $k \in \mathbb{R}$ is a hyperparameter. 

\paragraph{Training} Let $\by \in p_{\text{data}}$ be the one-hot representation of a label with $C$ classes and $\bs^{\by} = \{\pm k\}^{|C|}$ be its $k$-logit simplex representation of $\by$. The simplex diffusion model forward process $q(\bs^{\by}_t | \bs^{\by}_{t-1})$ is defined as a Gaussian-Markov process that produces a sequence of latent variables $\bs^{\by}_1, \dots, \bs^{\by}_T$ by gradually adding Gaussian noise at each time step $t \in {1, 2, \dots, T}$ with variance $\beta_t \in \mathbb{R}_{>0}$:

\begin{equation}
    q(\bs^{\by}_t | \bs^{\by}_{t-1}) = \mathcal{N}(\bs^{\by}_t | (1-\beta_t)\bs^{\by}_{t-1}, \beta_t \mathbf{I})
\label{eq2}
\end{equation}

Let $\boldsymbol{\epsilon}_t \sim \mathcal{N}(0, k^2\mathbf{I})$ as we convert data into simplex space, $\alpha_t = 1-\beta_t$, and $\bar{\alpha}_t=\prod^{t}_{j=1}\alpha_j$. Sampling $\bs^{\by}_t$ at an arbitrary time step $t$ has a closed-form solution:

\begin{equation}
    \bs^{\by}_t = \sqrt{\bar{\alpha}_t}\bs^{\by}_0 + \sqrt{1-\bar{\alpha}_t}\boldsymbol{\epsilon}_t
\label{eq3}
\end{equation}

Given a well-behaved noise schedule $\{\beta_t\}^T_{t=1}$, a little amount of Gaussian noise with variance $\beta_t$ is injected, while a large amount $1-\beta_t$ of previous sample $\bs^{\by}_{t-1}$ is preserved for each time step $t$. At the last time step $t=T$, our original data is expected to be no different from pure Gaussian distribution $\mathcal{N}(0, \mathbf{I})$. Therefore, in the denoising process, we can sample random noise from a standard Gaussian distribution and recover it sequentially to samples from $p_{\text{data}}$. Such an approximation of the reverse process $q(\bs^{\by}_{t-1}|\bs_t, \bs_0)$ can be delivered via a neural network with parameters $\boldsymbol{\theta}$, $p_{\boldsymbol{\theta}}(\bs^{\by}_{t-1}|\bs^{\by}_{t})$. In the context of our posterior estimation, the neural network is conditioned on $\bs^{\tilde{\by}}$, where $\tilde{\by}$ is the noisy label, to approximate $\bs^{\by}_{t-1}$ at time step $t$. The reverse process then is parameterized as

\begin{equation}
   \boldsymbol{p_{\theta}}(\bs^{\by}_{t-1}|\bs^{\by}_t, \bs^{\tilde{\by}}, \bx) = \mathcal{N}(\boldsymbol{\mu_{\theta}}(\bs^{\by}_t, t|\bs^{\tilde{\by}}, \bx), \boldsymbol{\Sigma_{\theta}}(\bs^{\by}_t, t|\bs^{\tilde{\by}}, \bx))
\label{eq4}
\end{equation}

As cross-entropy loss is typical in classification problems, we adopt it between the ground truth label and the model prediction given a noisy logit simplex $\bs_t$ at time step $t$.

\begin{equation}
   \mathcal{L} = \mathbb{L}_{t, q(\bs^{\by}_0 | \bs^{\tilde{\by}}, \bx_i), q(\bs^{\by}_t|\bs^{\by}_0, \bs^{\tilde{\by}}, \bx_i)}\Big[-\sum^{L}_{i=1}\log{\boldsymbol{p_{\theta}}(\by_i|\bs^{\by_i}_t, t, \bs^{\tilde{\by}_i}}, \bx_i) \Big]
\label{eq5}
\end{equation}

\paragraph{Noise Schedule} One important component in the diffusion forward process is the noise schedule. We follow the following cosine schedule for $\alpha_t$:

\begin{equation}
   \bar{\alpha}_t=\frac{f(t)}{f(0)}, \hspace{0.5cm} f(t)=\cos \Big(\frac{\frac{t}{T}+s}{1+s} \cdot \frac{\pi}{2} \Big)^2
\label{eq6}
\end{equation}

\paragraph{Inference} During inference of the simplex diffusion model,  $\bs_{T}$ is sampled from the prior $\mathcal{N}(0, k^2\mathbf{I})$. The model predictions are iteratively denoised for $t=T, \dots, 1$ starting from $k$-logit simplex Gaussian noise. This reverse process can be approximated via an adjustment of Equation (\ref{eq3}):

\begin{equation}
   \bs_{t-1} = \sqrt{\bar{\alpha}_{t-1}}\boldsymbol{\hat{S}_{\theta}}(\bs_t, t | \bs^{\tilde{\by}}, \bx) + \sqrt{1-\bar{\alpha}_{t-1}}\boldsymbol{\epsilon}_t
\label{eq7}
\end{equation}

where $\boldsymbol{\hat{S}_{\theta}}$ is the model prediction of the ground truth, $\bs^{\tilde{\by}}$ is noisy label simplex and $\bx$ is the input embedding, on which the model is conditioned. The model prediction $\boldsymbol{\hat{S}_{\theta}}(\bs_t, t | \bs^{\tilde{\by}}, \bx)$ is regarded as the hypothetical ground-truth and corrupts it by ($t-1$) time steps. To construct the model prediction, we project the logits produced by the underlying conditional model via argmax to match the initial $k$-logit representation:

\begin{equation}
    \hat{\bs}^{c}_{(i)}=
        \begin{cases} 
            k, &\text{if } i = \text{argmax(}\bs^{\by} \text{)}, \\
            -k &\text{otherwise.}
        \label{eq8}
        \end{cases}
\end{equation}

%% file: sections/04_exp.tex
\input{table/main_table}
\section{Experiments \& Results}
\label{experiments}
First, we introduce the tasks and datasets (20News Group, NumClaim, TREC, SemEval) that our experiments are conducted on (Section \ref{dataset}). Then, we describe our experimental setup (Section \ref{exp_setup}). Subsequently, we present the results of LLMs noise (Section \ref{llm_noise}), synthetic noise, and real-world noise (Section \ref{syn_real_noise}). Finally, we validate the effectiveness of each component in our framework (Section \ref{ablation_components}).

\subsection{Tasks and Datasets}
\label{dataset}
For our experiments, we include financial numerical claim detection from \citet{shah2024numerical}, question classification from \citet{li2002learning}, semantic relation classification task from \citet{hendrickx2019semeval}, and news topic modeling task from \citet{Lang95}. A summary of datasets used with the train-validation-test split is provided in table \ref{tb:dataset_summary}.

\begin{table}[H]
\footnotesize
\centering
\begin{tabular}{lccccc}
\toprule
Dataset   & Task         & \# Labels  & \multicolumn{3}{c}{Dataset Size}\\
            &    &             & Train & Valid & Test \\
\midrule
NumClaim   & Claim Detection      & 2 & 1715 & 429 & 537\\
TREC   & Question Classification & 6 & 5033 & 500 & 500\\
SemEval & Relation Extraction & 9 & 1749 & 178 & 600\\
20News & Topic Modeling & 20 & 9051 & 2263 & 7532\\
\bottomrule
\end{tabular}
\caption{Summary of datasets used. Dataset size denotes the number of samples in the benchmark.}
\label{tb:dataset_summary}
\end{table}
\vspace{-3.0em}
\subsection{Experimental Setup}
\label{exp_setup}
\paragraph{Baselines}
We compare SiDyP with the most relevant state-of-the-art baselines in the realm of learning from noisy labels. These baselines fall into three categories: (1) \textit{Basic Performances} without specific design tackling noisy labels \citep{devlin-etal-2019-bert}; (2) \textit{Multi-Model Training Strategies}: \textbf{Co-Teaching} \citep{han2018co} and \textbf{JoCoR} \citep{wei2020combating}. \textbf{Co-Teaching} trains two networks simultaneously and selects small-loss instances as clean samples for subsequent training. \textbf{JoCoR} also trains two networks simultaneously and uses co-regularization to achieve agreement to filter out noisy samples by selecting instances with small losses; (3) \textit{Generative Models for Noisy Maxtrix Estimation}: \textbf{NPC} \citep{bae2022noisy} and \textbf{DyGen} \citep{zhuang2023dygen}. \textbf{NPC} utilizes a generative model to calibrate the prediction of classifiers trained on noisy labels via a transition matrix. \textbf{DyGen} leverages the training dynamics to detect noisy samples and use a generative model to calibrate.

\input{table/llm_table}

\paragraph{Evaluation}
We use the model with the best validation accuracy during training for testing. We evaluate the methods by running on clean test datasets for all experiments. Given that the success of existing weakly-supervised learning methods relies heavily on clean validation samples \citep{zhu2023weakerthinkcriticallook}, we use noisy validation sets for model selections in all experiments. All experiments are run under 5 random seeds. We report the mean of the performances and the standard deviation.

\paragraph{Implementation Details}
We implement SiDyP using PyTorch \citep{paszke2019pytorchimperativestylehighperformance} and HuggingFace \citep{wolf2020huggingfacestransformersstateoftheartnatural}. We use BERT \citep{devlin-etal-2019-bert} classifier as our PLC in Stage I. Since random seeds affect network initialization, synthetic noise generation \citep{zhuang2023dygen, moschella2023relativerepresentationsenablezeroshot}, we use the same PLC results for the baselines which contain PLC fine-tuning on noisy label datasets (\textbf{NPC}, \textbf{DyGen}). More training details are revealed in Appendix \ref{ap:training_details}.

\subsection{LLMs Noise Experiments}
\label{llm_noise}
We run extensive experiments on various tasks and diversified LLM noises. First, we examine our framework in NumClaim, TREC, and SemEval labeled by \monospace{Llama-3-70b-chat-hf} \citep{llama3modelcard} in both zero-shot and few-shot manner. We use \monospace{Llama-3-70b-chat-hf} because it is the best open-source language model at the time when the experiments are conducted. We only prompt 20News Group in a zero-shot manner since it is a document-level task, and Llama-3-70b has a context length limitation of 8192, making it insufficient for few-shot learning. To test SiDyP under diversified LLM noises, we prompt \monospace{Meta-Llama-3.1-70B-Instruct-Turbo} \citep{llama3modelcard}, \monospace{Meta-Llama-3.1-405B-Instruct-Turbo} \citep{llama3modelcard}, \monospace{gpt-4o} \citep{openai2024gpt4technicalreport}, and \monospace{Mixtral-8x22B-Instruct-v0.1} \citep{jiang2024mixtral} in both zero-shot and few-shot prompting manners on SemEval task. In the following paragraphs, we present the experimental details and results.

\paragraph{LLM Prompting} For both zero-shot and few-shot manners, we use the same prompts for the same tasks for different LLMs (See prompting details in Appendix \ref{ap:prompts}). When the LLM is prompted to label the data, it is not guaranteed that it will follow the instructions and output the specified format. This leads to missing labels for some data samples in our annotated datasets. Although the portion of missing labels is trivial (i.e. highest missing label ratio 0.014\%  occurs in 20News Group dataset), we still preserve those data samples to maintain data's integrity for training and guarantee fair comparison among different LLMs. We assign a label to those missing-label samples uniformly. We use the dataset after random assignment for both training and validation. For the test dataset, we do not apply random assignment. The LLMs' raw accuracy is reported in Table \ref{tb:llama3-70b_result} and \ref{tb:LLMs_result}.

\paragraph{Results}
Table \ref{tb:llama3-70b_result} shows the results of Llama-3-70b on all four tasks. Our method (SiDyP) outperforms all baselines by a notable margin of 2.05\% across all tasks in both prompting manners. On average, there are 6.34\% samples of a fine-tuned PLC and 5.77\% of raw Llama-3-70b labeled samples successfully corrected by SiDyP. The performance gain on the SemEval task is the most significant, achieving an average increase of 3.7\%. This indicates that SiDyP is robust to the high noise ratio dataset. Although the base performance of NumClaim is competitive, SiDyP is able to bring an average of 20.19\% marginal increase. For NumClaim in a few-shot manner, our method is the only one to outperform Llama-3-70b raw labeling accuracy and fine-tuned PLC, demonstrating its effectiveness in the low noise ratio scenario.

\paragraph{Robustness Check for Diversified LLMs} Instead of only benchmarking Llama-3-70b, we extend our experiments to a variety of LLMs of different families with different sizes. We follow the same prompting and setting (See Appendix \ref{ap:llama_model} and \ref{ap:prompts}). We aim to check the robustness of our SiDyP framework under multiple LLM-generated label noise. Table \ref{tb:LLMs_result} shows the results of various types of LLM label noise on SemEval. Our method achieves significantly better performance compared to all baselines across all LLMs and both prompting manners. Specifically, SiDyP obtains an average performance gain of 4.47\% in comparison to the second-best baseline. Compared to a fine-tuned PLC on the noisy dataset, our method is able to boost the performance by an average of 8.02\%. In particular, a significant average increase of 11.73\% than LLMs raw accuracy is brought by our method. Combining all, we validate that our method is robust and resilient to different types of LLM noise and different prompting methods.

\input{table/syn_table}

\subsection{Synthetic and Real-world Noise Experiments}
\label{syn_real_noise}
Observing significant performance improvement in LLM-generated label noises, we further test our method under different families of noises, synthetic and real-world, on the SemEval task. We reveal the experiment details and results below.

\paragraph{Noise Generation} We inject three types of synthetic noises, including \textbf{Symmetric Noise (SN)}, \textbf{Asymmetric Noise (ASN)}, and \textbf{Instance-Dependent Noise (IDN)}. Symmetric Noise flips labels uniformly to other classes \citep{zhuang2023dygen, bae2022noisy, han2018co}. Asymmetric Noise flips labels with similar classes \citep{zhuang2023dygen, bae2022noisy}. Instance-Dependent Noise flips label with a probability proportional to the features of the sample \citep{zhuang2023dygen, bae2022noisy}. As synthetic noise is controllable, we use the noise ratio of 50\% to make a comparison with LLM noise. We choose 50\% because it aligns with the LLM noise ratio on SemEval. For real-world noise, we take the majority vote on the 164 labeling functions' output provided in WRENCH \citep{zhang2021wrench} for the SemEval dataset.

\paragraph{Results}
In Table \ref{tb:syn_real_result}, we present the results of various synthetic noises and real-world noises on SemEval. SiDyP achieves an average of 2.80\% increase compared to the second-best baseline. We observe that the performance increase between SiDyP and a strong baseline DyGen on LLM noises (5.21\%) is higher than it on synthetic noises (3.26\%). This is because DyGen performs better on synthetic datasets as such noises are less intricate \citep{zhuang2023dygen}. It further validates that LLM-generated label noises align more with real-world noise, making it more challenging for other baselines to arrive at accurate estimates. SiDyP, on the other hand, is resilient to different types of label noise and brings improvement consistently.

\input{table/ablation_table}
\begin{figure}[h]
    \includegraphics[width=0.9\linewidth]{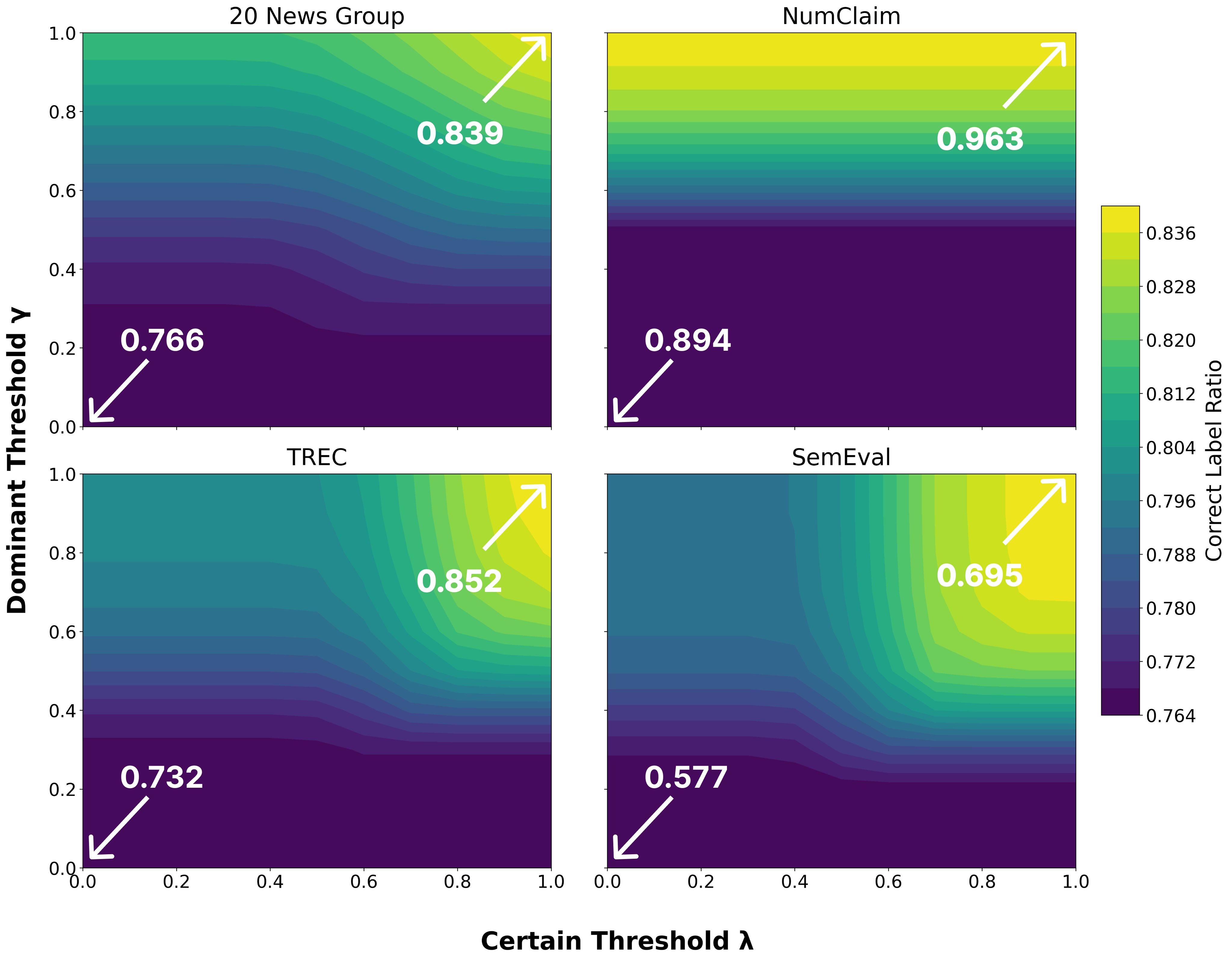}
    \caption{Label candidate correct ratio distribution across different combinations of certain threshold $\lambda$ and dominant threshold $\gamma$. Our label candidates acquire more correct labels for further generative label modeling. The arrows (to the left-bottom corner) point out the accuracy of fixed priors. The arrows (to the upper-bottom corner) point out the accuracy of dynamic priors, which is our method.}
    \label{fig:dynamic_prior}
\end{figure}
\vspace{-2em}
\subsection{Ablations}
\label{ablation_components}
\begin{figure}
    \includegraphics[width=\linewidth]{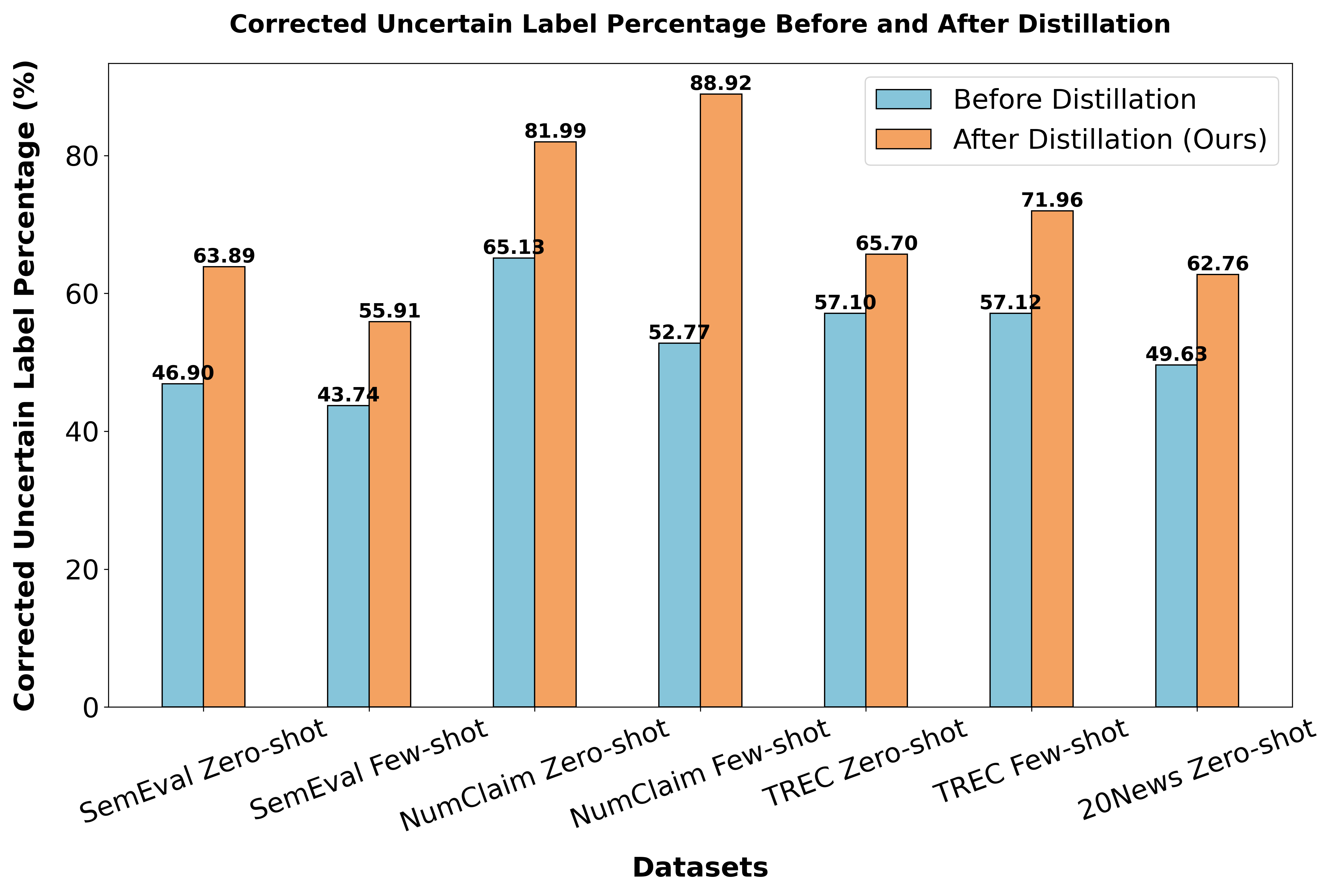}
    \caption{The percentage of uncertain labels being corrected by candidate distillation.}
    \label{fig:num_iter}
\end{figure}

To better understand the performance gain by SiDyP, we investigate the effectiveness of each component on Llama-3-70b labeled SemEval dataset in both zero-shot and few-shot manners. We eliminate them individually to validate their impact on performances in Table \ref{tb:ablation_result}: (1) Replacing our dynamic distillation priors with fix certain priors (for each sample, it's only associated with one fix certain label) in Stage II. (2) Substituting Stage II's simplex diffusion model with other generative models, Dirichlet variational auto-encoder (VAE) \citep{joo2019dirichletvariationalautoencoder} and Gaussian diffusion model \citep{sohldickstein2015deep,han2022card,chen2023labelretrievalaugmenteddiffusionmodelslearning}.
\paragraph{Label Candidate} We compare the portion of correct labels collected by our label candidate retrieval with the portion using the fix prior method to validate the improvement source of our dynamic prior. We calculate the accuracy of our label candidate for Llama-3-70b zero-shot labeled 20News Group, NumClaim, Trec, and SemEval across a wide range of certain threshold $\lambda$ and dominant threshold $\gamma$. For certain candidates, we directly compare it with the corresponding true label. For uncertain candidates, we either compare the specific candidate with maximum probability with the true label, or we check if the true label lies in our uncertain candidate. When $\lambda=\gamma=0$, the dynamic prior turns into a fix prior. Our label candidate achieves an average of 9.5\% improvement compared to the fix prior. Figure \ref{fig:dynamic_prior} presents the entire distribution of our dynamic prior accuracy.
\paragraph{Label Distillation} Figure \ref{fig:num_iter} presents the performance increase brought by our candidate dynamic distillation algorithm. We use all four datasets labeled by Llama-3-70B. We obtain the number of data instances that are corrected by Algorithm \ref{alg:distillation} in our training set. The corrected uncertain ratio is calculated by such an amount dividing the total number of uncertain data instances that contain true labels in their candidates. We are able to correct an average of 16.95\% samples across 7 noisy datasets, demonstrating the effectiveness of our distillation algorithm.
\paragraph{Generative Model}Our simplex denoising label diffusion model surpasses Dirichlet VAE by an average of 2.17\% and the Gaussian diffusion model by 8.58\%. Simplex diffusion directly models label probabilities within the simplex, ensuring every intermediate vector remains a valid distribution. Its iterative denoising process is especially effective in mapping noisy or partial label signals back to refined, true label distributions. By continually refining these probability vectors, it captures complex label distributions more reliably.

%% file: table/main_table.tex
\begin{table*}[ht]
\centering
\renewcommand{\arraystretch}{1.2} 
\begin{tabular}{lccccccc}
\toprule
\textbf{Datasets ($\rightarrow$)} & \multicolumn{2}{c}{\textbf{NumClaim}} & \multicolumn{2}{c}{\textbf{TREC}} & \multicolumn{2}{c}{\textbf{SemEval}} & \multicolumn{1}{c}{\textbf{20News}} \\
\cmidrule(lr){2-3} \cmidrule(lr){4-5} \cmidrule(lr){6-7} \cmidrule(lr){8-8} \textbf{Method ($\downarrow$)} & \textbf{Zero-shot} & \textbf{Few-shot} & \textbf{Zero-shot} & \textbf{Few-shot} & \textbf{Zero-shot} & \textbf{Few-shot} & \textbf{Zero-shot} \\
\midrule
Llama-3-70b     & 89.94 & 95.53 & 81.80 & 84.00 & 47.50 & 48.50 & 74.04 \\
PLC             & 90.54\scriptsize{$\pm$0.72} & 95.11\scriptsize{$\pm$0.30} & 80.64\scriptsize{$\pm$0.94} & 77.72\scriptsize{$\pm$1.34} & 51.59\scriptsize{$\pm$0.44} & 50.46\scriptsize{$\pm$0.72} & 71.2\scriptsize{$\pm$0.52} \\
Co-teaching     & 92.25\scriptsize{$\pm$1.17} & 94.45\scriptsize{$\pm$0.53} & 80.24\scriptsize{$\pm$3.51} & 80.42\scriptsize{$\pm$2.55} & 52.45\scriptsize{$\pm$1.43} & 50.90\scriptsize{$\pm$1.36} & 70.95\scriptsize{$\pm$0.64} \\
JoCoR           & 92.14\scriptsize{$\pm$0.61} & 93.45\scriptsize{$\pm$0.80} & 82.22\scriptsize{$\pm$0.89} & 82.08\scriptsize{$\pm$0.39} & 53.30\scriptsize{$\pm$1.11} & 53.60\scriptsize{$\pm$0.68} & 70.95\scriptsize{$\pm$0.63} \\
NPC             & 90.83\scriptsize{$\pm$0.62} & 95.04\scriptsize{$\pm$0.61} & 79.48\scriptsize{$\pm$1.97} & 78.88\scriptsize{$\pm$1.47} & 50.73\scriptsize{$\pm$1.70} & 47.53\scriptsize{$\pm$1.26} & 70.60\scriptsize{$\pm$0.51} \\
DyGen           & 91.13\scriptsize{$\pm$0.30} & 95.41\scriptsize{$\pm$0.28} & 82.88\scriptsize{$\pm$0.71} & 84.80\scriptsize{$\pm$0.86} & 60.86\scriptsize{$\pm$0.81} & 60.79\scriptsize{$\pm$2.23} & 71.42\scriptsize{$\pm$0.31} \\
\rowcolor{rowblue}
\textbf{SiDyP}  & \textbf{93.63\scriptsize{$\pm$0.84}} & \textbf{95.97\scriptsize{$\pm$0.15}} & \textbf{84.76\scriptsize{$\pm$0.79}} & \textbf{85.60\scriptsize{$\pm$0.44}} & \textbf{64.26\scriptsize{$\pm$0.27}} & \textbf{64.79\scriptsize{$\pm$0.96}} & \textbf{72.66\scriptsize{$\pm$0.58}} \\
\bottomrule
\end{tabular}
\caption{Results on Llama-3-70b noise. Numbers reported are classification accuracy. \textbf{Bold} represents the best performance.}
\label{tb:llama3-70b_result}
\end{table*}

%% file: table/llm_table.tex
\begin{table*}[h]
\centering
\renewcommand{\arraystretch}{1.2} 
\begin{tabular}{lcccccccc}
\toprule
\textbf{Dataset} ($\rightarrow$) & \multicolumn{8}{c}{\textbf{SemEval}} \\
\midrule
\multirow{2}{*}{\textbf{Method ($\downarrow$)}} & \multicolumn{2}{c}{\textbf{Llama-3.1-70b}} & \multicolumn{2}{c}{\textbf{Llama-3.1-405b}} & \multicolumn{2}{c}{\textbf{GPT4o}} & \multicolumn{2}{c}{\textbf{Mixtral-8x22b}} \\
\cmidrule(lr){2-3} \cmidrule(lr){4-5} \cmidrule(lr){6-7} \cmidrule(lr){8-9}
 & \textbf{Zero-shot} & \textbf{Few-shot} & \textbf{Zero-shot} & \textbf{Few-shot} & \textbf{Zero-shot} & \textbf{Few-shot} & \textbf{Zero-shot} & \textbf{Few-shot} \\
\midrule
Base            & 52.66 & 55.16 & 55.16 & 52.16 & 56.50 & 57.66 & 42.66 & 40.83 \\
PLC             & 60.26\scriptsize{$\pm$0.89} & 57.70\scriptsize{$\pm$1.10} & 54.76\scriptsize{$\pm$1.24} & 53.96\scriptsize{$\pm$0.12} & 58.63\scriptsize{$\pm$0.86} & 61.56\scriptsize{$\pm$0.93} & 49.29\scriptsize{$\pm$1.31} & 46.33\scriptsize{$\pm$1.32} \\
Co-teaching     & 59.08\scriptsize{$\pm$2.92} & 58.52\scriptsize{$\pm$0.30} & 55.27\scriptsize{$\pm$1.43} & 54.92\scriptsize{$\pm$2.87} & 62.82\scriptsize{$\pm$1.19} & 65.75\scriptsize{$\pm$1.17} & 50.48\scriptsize{$\pm$2.66} & 46.47\scriptsize{$\pm$1.10} \\
JoCoR           & 63.38\scriptsize{$\pm$3.56} & 61.22\scriptsize{$\pm$1.16} & 57.28\scriptsize{$\pm$0.92} & 56.96\scriptsize{$\pm$1.09} & 62.13\scriptsize{$\pm$1.27} & 64.78\scriptsize{$\pm$1.18} & 51.45\scriptsize{$\pm$1.42} & 45.45\scriptsize{$\pm$2.94} \\
NPC             & 60.13\scriptsize{$\pm$0.77} & 57.49\scriptsize{$\pm$3.00} & 55.06\scriptsize{$\pm$2.99} & 54.53\scriptsize{$\pm$1.24} & 59.56\scriptsize{$\pm$0.90} & 61.40\scriptsize{$\pm$1.53} & 47.56\scriptsize{$\pm$1.26} & 41.96\scriptsize{$\pm$0.70} \\
DyGen           & 68.53\scriptsize{$\pm$0.88} & 64.53\scriptsize{$\pm$2.85} & 59.69\scriptsize{$\pm$1.31} & 51.69\scriptsize{$\pm$2.02} & 62.63\scriptsize{$\pm$0.91} & 64.03\scriptsize{$\pm$0.82} & 50.63\scriptsize{$\pm$6.43} & 40.23\scriptsize{$\pm$1.41} \\
\rowcolor{rowblue}
\textbf{SiDyP}  & \textbf{71.66\scriptsize{$\pm$0.91}} & \textbf{67.43\scriptsize{$\pm$1.36}} & \textbf{62.76\scriptsize{$\pm$0.99}} & \textbf{60.46\scriptsize{$\pm$2.06}} & \textbf{66.86\scriptsize{$\pm$0.48}} & \textbf{68.83\scriptsize{$\pm$1.07}} & \textbf{57.96\scriptsize{$\pm$1.94}} & \textbf{50.66\scriptsize{$\pm$2.02}} \\
\bottomrule
\end{tabular}
\caption{Results on four different LLMs noises. "Base" represents LLM's raw accuracy on test sets.}
\label{tb:LLMs_result}
\end{table*}

%% file: table/syn_table.tex
\begin{table}[H]
\centering
\renewcommand{\arraystretch}{1.2} 
\resizebox{0.48\textwidth}{!}{
\begin{tabular}{lcccc}
\toprule
\textbf{Datasets ($\rightarrow$)} & \multicolumn{4}{c}{\textbf{SemEval}} \\
\cmidrule(lr){2-5} \textbf{Method ($\downarrow$)} & \textbf{SN} & \textbf{ASN} & \textbf{IDN} & \textbf{Real World} \\
\midrule
Base            & 50.00 & 50.00 & 50.00 & 82.50\\
PLC             & 65.06\scriptsize{$\pm$2.13} & 40.96\scriptsize{$\pm$2.60} & 59.83\scriptsize{$\pm$2.65} & 84.13\scriptsize{$\pm$0.68} \\
Co-teaching     & 70.27\scriptsize{$\pm$1.77} & 43.90\scriptsize{$\pm$7.39} & 67.13\scriptsize{$\pm$4.38} & 78.98\scriptsize{$\pm$1.05} \\
JoCoR           & 73.45\scriptsize{$\pm$1.58} & 44.92\scriptsize{$\pm$7.78} & 71.05\scriptsize{$\pm$5.51} & 79.22\scriptsize{$\pm$1.43} \\
NPC             & 57.73\scriptsize{$\pm$3.61} & 42.60\scriptsize{$\pm$5.46} & 54.16\scriptsize{$\pm$4.91} & 81.23\scriptsize{$\pm$1.88} \\
DyGen           & 73.06\scriptsize{$\pm$2.07} & 53.16\scriptsize{$\pm$5.46} & 71.40\scriptsize{$\pm$1.80} & 82.30\scriptsize{$\pm$0.13} \\
\rowcolor{rowblue}
\textbf{SiDyP}  & \textbf{74.26\scriptsize{$\pm$1.99}} & \textbf{59.63\scriptsize{$\pm$3.06}} & \textbf{73.19\scriptsize{$\pm$2.22}} & \textbf{85.86\scriptsize{$\pm$0.52}} \\
\bottomrule
\end{tabular}}
\caption{Performance comparison on SemEval with synthetic noise (SN, ASN, IDN) and real-world noise.}
\label{tb:syn_real_result}
\end{table}

%% file: table/ablation_table.tex
\begin{table}[H]
\centering
\renewcommand{\arraystretch}{1.2} 
\begin{tabular}{lcc}
\toprule
\textbf{Datasets ($\rightarrow$)} & \multicolumn{2}{c}{\textbf{SemEval}} \\
\cmidrule(lr){2-3} \textbf{Method ($\downarrow$)} & \textbf{Zero-shot} & \textbf{Few-shot}\\
\midrule
FP + Dir-VAE                       & 60.86\scriptsize{$\pm$0.81} & 60.79\scriptsize{$\pm$2.23}\\
FP + Sim-Diff                      & 62.73\scriptsize{$\pm$1.06} & 63.26\scriptsize{$\pm$1.06} \\
DP + Gau-Diff                      & 54.53\scriptsize{$\pm$3.48} & 57.36\scriptsize{$\pm$3.64} \\
\rowcolor{rowblue}
DP + Sim-Diff (\textbf{SiDyP})     & \textbf{64.26\scriptsize{$\pm$0.27}} & \textbf{64.79\scriptsize{$\pm$0.96}} \\
\bottomrule
\end{tabular}
\caption{Different components efficacy on zero-shot and few-shot labeled SemEval by Llama-3-70b. "FP"=fix prior. "DP"=our dynamic prior. "Dir-VAE"=Dirichlet VAE. "Gau-Diff"=Gaussian diffusion model. "Sim-Diff"=simplex diffusion model.} 
\label{tb:ablation_result}
\end{table}

%% file: sections/05_relate.tex
\section{Related Work}
\label{prework}
\paragraph{Weak Supervision} Weak supervision in machine learning includes incomplete, inexact, and inaccurate categories, each tailored to specific imperfections in data \citep{zhou2018brief}. Inexact supervision deals with broad labels, while inaccurate supervision, where labels are erroneous, employs techniques like data programming \citep{ratner2017snorkel}, human-in-the-loop strategies \citep{zhang2022prboost}, and contrastive loss for enhanced learning from data similarities and differences \citep{yu2020fine}. \citet{zhang2021wrench} apply a two-stage model to manage inaccurate supervision, initially denoising data before training on refined labels.
\paragraph{LLM as annotators} LLMs have also been leveraged to iteratively expand label space under extremely weak supervision. ChatGPT is verified as a more accurate and cost-effective method than traditional scalable crowdsourcing \citep{Gilardi_2023}. X-MLClass \citep{li2024openworldmultilabeltextclassification} demonstrated significant improvements in label discovery and multi-label classification accuracy in open-world settings. Additionally, explanation-aware ensembling methods like EASE \citep{yu2023explanation} further illustrate how LLMs can be used to improve in-context learning by effectively guiding predictions and mitigating label noise. 
\paragraph{Learning from Noisy Labels}In the landscape of learning from noisy labels, \citet{iscen2022learning} proposed that there are similarities among training instances in the feature/embedding space, leading to the consistency of labels between data instances and their neighbors. NPC \citep{bae2022noisy} lies in the class of transition matrix base method. The true label is inferred by a prior, estimated by a pre-trained classifier, and a posterior, approximated by a generative model. DyGen \citep{zhuang2023dygen} infers a true label based on the training dynamics during the finetuning of the pre-trained language model. The feasibility of Diffusion Models in classification problems is explored and validated by \cite{han2022card}. \citet{chen2023labelretrievalaugmented} is the very first to exploit the Gaussian diffusion model in the context of noisy label learning. \citep{wang2023noiserobustfinetuningpretrainedlanguage} utilizes LLMs as an external guider to distinguish clean and noisy samples. Although the problem of learning from noisy labels is studied sophistically, research on enhancing learning from LLM-generated label noises is lagging, yet urgently needed.

%% file: sections/06_conclusion.tex
\section{Conclusion}
\label{conclusion}
In this paper, we highlight the importance of improving learning from LLM-generated noisy labels. The emergence of LLMs has provided a cost-effective alternative to traditional data annotation methods, yet the presence of noisy labels remains a critical challenge. We propose a denoising framework SiDyP, effectively mitigating the impact of LLM-generated noisy labels by leveraging neighborhood label distribution in embedding space and refining label predictions through a simplex diffusion model. Experimental results demonstrate that SiDyP significantly enhances classifier performance, achieving an average improvement of 7.21\% and 7.30\% on zero-shot and few-shot LLM-generated noisy datasets, respectively. By benchmarking across multiple LLMs and NLP tasks, we highlight the limitations of existing noisy label learning approaches and establish SiDyP as a robust denoising method. Our findings open new directions for research in learning from LLM-generated label noise.

%% file: sections/appendix.tex

\appendix
\section{Dataset and Task Detail}
\label{ap:dataset_task_detail}
\begin{itemize}
    \item \textbf{Numerical Claim Detection (NumClaim)}: This involves extracting numerical claims from financial texts like analysts' reports to forecast stock price volatility. Using a dataset with binary labels for sentences, this task distinguishes between "in-claim" sentences that predict financial outcomes and "out-of-claim" sentences that state factual information.

    \item \textbf{Question Classification (TREC)}: This task involves classifying questions into predefined categories based on their intent and content, as outlined in the TREC dataset from \citet{li2002learning} study. Using a dataset of labeled questions, this task assigns each question to one of six categories: location, entity, description, human, numeric value, and abbreviation. The goal is to determine the type of answer each question seeks, thereby facilitating targeted information retrieval and enhancing the efficiency of question-answering systems.

    \item \textbf{Semantic Relation Extraction (SemEval)}: This task focuses on the multi-way classification of semantic relations between pairs of nominals, as defined in SemEval-2010 Task 8 \citep{hendrickx2019semeval}. Utilizing a dataset where each pair of nominals is annotated with one of nine (Cause-Effect, Instrument-Agency, etc.) possible semantic relations, this task involves determining the specific type of relationship that exists between the two terms. The nine categories include Cause-Effect, Instrument-Agency, Product-Producer, Content-Container, Entity-Origin, Entity-Destination, Component-Whole, Member-Collection, and Message-Topic. The objective is to enhance the understanding of linguistic patterns and to improve the semantic analysis capabilities of natural language processing systems.
    \item \textbf{News Topic Modeling (20News)}: This task involves classifying news articles into different topics using the well-known 20 Newsgroups dataset \citep{Lang95}. The dataset contains around 20,000 documents collected from newsgroups, organized into 20 different categories such as 'rec.sport.baseball', 'comp.graphics', and 'sci.med'. Each document is assigned to one of these categories. The task's objective is to train models to effectively capture the topical structure of news articles, which helps improve text categorization and topic detection capabilities in natural language processing applications. 
\end{itemize}
\section{LLM Prompting Details}
\subsection{Model Implementation Details}
\label{ap:llama_model}
We take advantage of API from \href{https://www.together.ai/}{Together.ai}. We are grateful to them for providing free credits and making it possible. We use the model with a $temperature$ value of 0.00 (for reproducibility) and $max\_token$ of 100.

\subsection{Prompt Templates}
\label{ap:prompts}
\begin{tcolorbox}[colback=gray!10, colframe=black, title=\textbf{NumClaim Zero-shot Prompt Template}]
    \textbf{Prompt} \\
    \textbf{USER:} Classify the following sentence into 'INCLAIM', or 'OUTOFCLAIM' class. 'INCLAIM' refers to predictions or expectations about financial outcomes, it can be thought of as 'financial forecasts'. 'OUTOFCLAIM' refers to sentences that provide numerical information or established facts about past financial events. Now, for the following sentence provide the label in the first line and provide a short explanation in the second line. The sentence: \textcolor{blue}{\monospace{\{sentence\}}} \\
\end{tcolorbox}

\begin{tcolorbox}[colback=gray!10, colframe=black, title=\textbf{TREC Zero-shot Prompt Template}]
    \textbf{Prompt} \\
    \textbf{USER:} For the following question, which belongs to a specific category, categorize it into one of the following classes based on the type of answer it requires: Abbreviation (ABBR), Entity (ENTY), Description (DESC), Human (HUM), Location (LOC), Numeric (NUM). Provide the label in the first line and provide a short explanation in the second line. The question: \textcolor{blue}{\monospace{\{question\}}}
\end{tcolorbox}

\begin{tcolorbox}[colback=gray!10, colframe=black, title=\textbf{SemEval Zero-shot Prompt Template}]
    \textbf{Prompt} \\
    \textbf{USER:} The task is to identify the type of semantic relationship between two nominals in a given sentence. Below are the definitions of the nine relationship categories you must choose from:\\
    Cause-Effect (CE): An event or object leads to an effect.\\
    Instrument-Agency (IA): An agent uses an instrument.\\
    Product-Producer (PP): A producer causes a product to exist.\\
    Content-Container (CC): An object is physically stored in a delineated area of space.\\
    Entity-Origin (EO): An entity is coming or is derived from an origin (e.g., position or material).\\
    Entity-Destination (ED): An entity is moving towards a destination.\\
    Component-Whole (CW): An object is a component of a larger whole.\\
    Member-Collection (MC): A member forms a nonfunctional part of a collection.\\
    Message-Topic (MT): A message, written or spoken, is about a topic.\\
    For the provided sentence below, determine the most accurate relationship category based on the descriptions provided. Respond by selecting the label (e.g., CE, IA, PP, etc.) that best matches the relationship expressed in the sentence. Provide the label in the first line and provide a short explanation in the second line. The sentence: \textcolor{blue}{\monospace{\{sentence\}}}
\end{tcolorbox}

\begin{tcolorbox}[colback=gray!10, colframe=black, title=\textbf{NumClaim Few-shot Prompt Template}]
    \textbf{Prompt} \\
    \textbf{USER:} Classify the following sentence into 'INCLAIM', or 'OUTOFCLAIM' class. 'INCLAIM' refers to predictions or expectations about financial outcomes, it can be thought of as 'financial forecasts'. 'OUTOFCLAIM' refers to sentences that provide numerical information or established facts about past financial events. Here are two examples:\\
    Example 1:  \textcolor{blue}{\monospace{\{example\}}} // OUTOFCLAIM \\
    Example 2: \textcolor{blue}{\monospace{\{example\}}} // INCLAIM \\
    Now, for the following sentence provide the label in the first line and provide a short explanation in the second line. The sentence: \textcolor{blue}{\monospace{\{sentence\}}}
\end{tcolorbox}

\begin{tcolorbox}[colback=gray!10, colframe=black, title=\textbf{TREC Few-shot Prompt Template}]
    \textbf{Prompt} \\
    \textbf{USER:} For the following question, which belongs to a specific category, categorize it into one of the following classes based on the type of answer it requires: Abbreviation (ABBR), Entity (ENTY), Description (DESC), Human (HUM), Location (LOC), Numeric (NUM). Here are six examples:\\
    Example 1: \textcolor{blue}{\monospace{\{example\}}} // DESC\\
    Example 2: \textcolor{blue}{\monospace{\{example\}}} // ENTY\\
    Example 3: \textcolor{blue}{\monospace{\{example\}}} // HUM\\
    Example 4: \textcolor{blue}{\monospace{\{example\}}} // ABBR\\
    Example 5: \textcolor{blue}{\monospace{\{example\}}} // LOC\\
    Example 6: \textcolor{blue}{\monospace{\{example\}}} // NUM \\
    Now for the following question provide the label in the first line and provide a short explanation in the second line. The question: \textcolor{blue}{\monospace{\{question\}}}
\end{tcolorbox}

\begin{tcolorbox}[colback=gray!10, colframe=black, title=\textbf{SemEval Few-shot Prompt Template}]
    \textbf{Prompt} \\
    \textbf{USER:} The task is to identify the type of semantic relationship between two nominals in a given sentence. Below are the definitions of the nine relationship categories you must choose from:\\
    Cause-Effect (CE): An event or object leads to an effect. (Example:\textcolor{blue}{\monospace{\{example\}}})\\
    Instrument-Agency (IA): An agent uses an instrument. (Example:\textcolor{blue}{\monospace{\{example\}}})\\
    Product-Producer (PP): A producer causes a product to exist. (Example:\textcolor{blue}{\monospace{\{example\}}})\\
    Content-Container (CC): An object is physically stored in a delineated area of space. (Example:\textcolor{blue}{\monospace{\{example\}}})\\
    Entity-Origin (EO): An entity is coming or is derived from an origin (e.g., position or material) (Example:\textcolor{blue}{\monospace{\{example\}}}).\\
    Entity-Destination (ED): An entity is moving towards a destination. (Example:\textcolor{blue}{\monospace{\{example\}}})\\
    Component-Whole (CW): An object is a component of a larger whole. (Example:\textcolor{blue}{\monospace{\{example\}}})\\
    Member-Collection (MC): A member forms a nonfunctional part of a collection. (Example:...)\\
    Message-Topic (MT): A message, written or spoken, is about a topic. (Example:\textcolor{blue}{\monospace{\{example\}}})\\
    For the provided sentence below, determine the most accurate relationship category based on the descriptions provided. Respond by selecting the label (e.g., CE, IA, PP, etc.) that best matches the relationship expressed in the sentence. Provide the label in the first line and provide a short explanation in the second line. The sentence: \textcolor{blue}{\monospace{\{sentence\}}}
\end{tcolorbox}

\begin{tcolorbox}[colback=gray!10, colframe=black, title=\textbf{20News Group Zero-shot Prompt Template}]
    \textbf{Prompt} \\
    \textbf{USER:} The task is to classify the given text into one of the 20 news group categories. Below are the 20 categories you must choose from:\\1. 'alt.atheism': Discussions related to atheism.\\2. 'comp.graphics': Topics about computer graphics, including software and hardware.\\3. 'comp.os.ms-windows.misc': Discussions about the Microsoft Windows operating system.\\4. 'comp.sys.ibm.pc.hardware': Topics related to IBM PC hardware.\\5. 'comp.sys.mac.hardware': Discussions about Mac hardware.\\6. 'comp.windows.x': Topics about the X Window System.\\7. 'misc.forsale': Posts related to buying and selling items.\\8. 'rec.autos': Discussions about automobiles.\\9. 'rec.motorcycles': Topics related to motorcycles.\\10. 'rec.sport.baseball': Discussions about baseball.\\11. 'rec.sport.hockey': Discussions about hockey.\\12. 'sci.crypt': Topics about cryptography and encryption.\\13. 'sci.electronics': Discussions about electronic systems and devices.\\14. 'sci.med': Topics related to medical science and healthcare.\\15. 'sci.space': Discussions about space and astronomy.\\16. 'soc.religion.christian': Topics about Christianity and related discussions.\\17. 'talk.politics.guns': Discussions about gun politics and related debates.\\18. 'talk.politics.mideast': Topics about politics in the Middle East.\\19. 'talk.politics.misc': General political discussions not covered by other categories.\\20. 'talk.religion.misc': Discussions about miscellaneous religious topics.\\For the provided text below, determine the most appropriate category based on the descriptions above. Respond by selecting the label (e.g., alt.atheism, comp.graphics, etc.) that best matches the topic of the text. Provide the label in the first line and a brief explanation in the second line. The sentence: \textcolor{blue}{\monospace{\{sentence\}}}
\end{tcolorbox}

\section{Training Dynamics and Co-Regularization}
\label{td_cr}
\paragraph{Training Dynamics} The training dynamics during PLC fine-tuning (Stage I in Figure \ref{fig:sidyp_pipeline}) is not only beneficial for clean and noisy sample separation (as we discuss in Section \ref{dy_distill}), but also contains rich information attributing to generative model learning (Stage II in Figure \ref{fig:sidyp_pipeline}) \citep{zhuang2023dygen}. Leveraging such dynamics, our empirical objective becomes:
$$p(y|x) \propto \sum_{\hat{y}}p(\hat{y} | x)p(y|\hat{y}, W)$$
where $W$ denotes the training dynamics for each sample.
\balance
\paragraph{Co-Regularization} Although we manage to mitigate the negative impact of label noises (Section \ref{dy_distill},\ref{simplex_diff}), it is inevitable that small deviations in $p(\hat{y}|x)$ and $p(y|\hat{y},x)$ could propagate to later stages, thus affecting the objective $p(y|x)$. We leverage multiple branches with identical architecture but different initializations \citep{zhuang2023dygen}. A co-regularization loss across branches is introduced to achieve consensus. Such a loss is calculated as the KL Divergence between the consensus probability (the average probability of models' predicted probability in different model branches) and each individual model's predicted probability. We apply co-regularization mechanism to both Stage I PLC $\mathbf{F}_{\varphi}(\hat{y}|x)$ and Stage II generative model $p_{\theta}(y|\hat{y},x)$. To begin, we initialize $M$ copies of $\mathbf{F}^{(m)}_{\varphi}(\hat{y}|x)$ and $p^{(m)}_{\theta}(y|\hat{y},x)$. Passing instances $x_i$ to different model branches, we can obtain the corresponding model predicted probabilities $p^{(m)}_i$. Then, an aggregated probability $q_i$ can be calculated by averaging all predicted probabilities:
$$q_i = \frac{1}{M}\sum^M_{m=1}p^{(m)}_i$$
Given these, a co-regularization loss can be calculated as follows:
\begin{align*}
    \ell_{\text{CR}}&=\frac{1}{MN}\sum^N_{i=1}\sum^M_{m=1}\text{KLK}(q_i || p^{(m)}_i) \\
    &= \frac{1}{MN}\sum^N_{i=1}\sum^M_{m=1}\sum^C_{c=1}q_{ic}\log \Big( \frac{q_{ic}+\epsilon}{p^{(m)}_{ic}+\epsilon}\Big)
\end{align*}
where $\epsilon$ indicates a small positive number to avoid division by zero.

\begin{table*}[ht]
\centering
\renewcommand{\arraystretch}{1.2} 
\begin{tabular}{lccccccc}
\toprule
\textbf{LLM} ($\rightarrow$) & \multicolumn{7}{c}{\textbf{Llama-3-70b}} \\
\midrule
\textbf{Datasets ($\rightarrow$)} & \multicolumn{2}{c}{\textbf{NumClaim}} & \multicolumn{2}{c}{\textbf{TREC}} & \multicolumn{2}{c}{\textbf{SemEval}} & \multicolumn{1}{c}{\textbf{20News}} \\
\cmidrule(lr){2-3} \cmidrule(lr){4-5} \cmidrule(lr){6-7} \cmidrule(lr){8-8} \textbf{Method ($\downarrow$)} & \textbf{Zero-shot} & \textbf{Few-shot} & \textbf{Zero-shot} & \textbf{Few-shot} & \textbf{Zero-shot} & \textbf{Few-shot} & \textbf{Zero-shot} \\
\midrule
$E_{\text{BERT}}$ & 10 & 10 & 10 & 10 & 10 & 10 & 10 \\
batch size  & 128 & 128 & 128 & 128 & 128 & 128 & 128 \\
learning rate (BERT) & 5e-5 & 5e-5 & 5e-5 & 5e-5 & 5e-5 & 5e-5 & 5e-5 \\
max length  & 128 & 128 & 128 & 128 & 128 & 128 & 128 \\
$\sigma$ & 0.1 & 0.05 & 0.3 & 0.3 & 0.5 & 0.5 & 0.3\\
$\lambda$ & 0.9 & 0.8 & 0.8 & 0.8 & 0.9 & 0.9 & 0.8 \\
$\gamma$ & 0.9 & 0.8 & 0.8 & 0.8 & 0.8 & 0.8  & 0.8\\
$\alpha$ & 4 & 2 & 1 & 2 & 2 & 3 & 2\\
$m$ & 3 & 3 & 3 & 3 & 3 & 3 & 3\\
$\beta$ & 4 & 4 & 4 & 4 & 2 & 4  & 6 \\
$E_{\text{SD}}$ & 10 & 10 & 10 & 10 & 10 & 10 & 10 \\
batch size (SD) & 128 & 128 & 128 & 128 & 128 & 128 & 128\\
learning rate (SD) & 5e-4 & 5e-4 & 5e-4 & 6e-4 & 5e-4 & 1e-4 & 5e-4\\
train timesteps & 500 & 200 & 500 & 200 & 800 & 500 & 500\\
inference timesteps & 10 & 10 & 100 & 10 & 10 & 100 & 100  \\
K & 10 & 20 & 10 & 20 & 10 & 10 & 10  \\
\bottomrule
\end{tabular}
\caption{Training hyper-parameters details for SiDyP on all six Llama-3 generated datasets.} 
\label{tb:hyperparameter}
\end{table*}

\begin{table*}[ht]
\renewcommand{\arraystretch}{1.2} 
\setlength{\tabcolsep}{3.3pt} 
\begin{tabular}{lccccccc}
\toprule
\textbf{LLM} ($\rightarrow$) & \multicolumn{7}{c}{\textbf{Llama-3-70b}} \\
\midrule
\textbf{Datasets ($\rightarrow$)} & \multicolumn{2}{c}{\textbf{NumClaim}} & \multicolumn{2}{c}{\textbf{TREC}} & \multicolumn{2}{c}{\textbf{SemEval}} & \multicolumn{1}{c}{\textbf{20News}} \\
\cmidrule(lr){2-3} \cmidrule(lr){4-5} \cmidrule(lr){6-7} \cmidrule(lr){8-8} \textbf{Method ($\downarrow$)} & \textbf{Zero-shot} & \textbf{Few-shot} & \textbf{Zero-shot} & \textbf{Few-shot} & \textbf{Zero-shot} & \textbf{Few-shot} & \textbf{Zero-shot} \\
\midrule
Noise Ratio (Original)     & 91.69 & 95.85 & 70.35 & 69.72 & 50.96 & 50.64 & 76.13  \\
No Answer Ratio            & 0.00 & 0.00 & 3.6$e^{-4}$ & 1.8$e^{-4}$ & 2.5$e^{-3}$ & 4.1$e^{-3}$ & 1.4$e^{-2}$  \\
Noise Ratio (After RA)     & 91.69 & 95.85 & 70.35 & 69.72 & 50.96 & 50.64 & 76.23  \\
\bottomrule
\end{tabular}
\caption{Llama-3-70b label noise ratio on training sets of 20News, NumClaim, TREC, and SemEval. "RA":random assignment.}
\label{tb:llama3-70b_noise_ratio}
\end{table*}

\section{SiDyP Training Details}
\label{ap:training_details}
 All experiments are conducted on CPU: Intel(R) Xeon(R) W-2295 CPU @ 3.00GHz and GPU: NVIDIA GeForce RTX A6000 GPUs using Python 3.11.5 and PyTorch 2.0.1. We use Adam \citep{kingma2017adam} as the optimizer. $E_{\text{BERT}}$ is the training epochs for the BERT classifier. $E_{\text{SD}}$ is the training epochs for the simplex diffusion model. $\sigma$ is the estimated error rate in Algorithm \ref{alg:retrieval}. $\lambda$ is the threshold that we separate certain and uncertain prior in Algorithm \ref{alg:retrieval}. $\gamma$ is the threshold that we preserve the dominance candidates in uncertain prior in Algorithm \ref{alg:retrieval}. In Algorithm \ref{alg:distillation}, $\alpha$ is the warmup epochs for Stage II generative model training. $m$ is the number of model branches. $\beta$ is the number of sample times that we use to refine our uncertain prior based on the model's predictions.
 \paragraph{Time Complexity} We perform Big-O analysis for SiDyP. The time complexity for SiDyP is $O(W^2 \times T)$ where $W$ denotes the embedding size of training dynamics and $T$ is either training timesteps or inference timesteps of our simplex diffusion model. We choose $\gamma$ based on our empirical estimation. To make a fair comparison, we use the same estimated error rate in all other baselines, which require one. We grid search these hyper-parameters: $\lambda$ in [0.7, 0.8, 0.9, 1.0], $\gamma$ in [0.4, 0.6, 0.8], $\alpha$ in [1, 2, 3, 4, 5, 6], $\beta$ in [2, 4, 6, 8], $K$ in [10, 20, 30], train timesteps in [400, 500, 600, 700, 800], inference timesteps in [10, 20, 50, 100], learning rate in [1e-3, 6e-4, 3e-4, 1e-4].

\section{LLM Noise Ratio}
\label{ap:llm_noise_ratio}
We present noise ratio of LLMs labeled training dataset in Table \ref{tb:llama3-70b_noise_ratio}, \ref{tb:LLMs_noise_ratio}. 

\begin{table*}[ht]
\centering
\renewcommand{\arraystretch}{1.2} 
\setlength{\tabcolsep}{1.3pt} 
\begin{tabular}{lcccccccc}
\toprule
\textbf{Dataset} ($\rightarrow$) & \multicolumn{8}{c}{\textbf{SemEval}} \\
\midrule
\multirow{2}{*}{\textbf{Method ($\downarrow$)}} & \multicolumn{2}{c}{\textbf{Llama-3.1-70b}} & \multicolumn{2}{c}{\textbf{Llama-3.1-405b}} & \multicolumn{2}{c}{\textbf{GPT4o}} & \multicolumn{2}{c}{\textbf{Mixtral-8x22b}} \\
\cmidrule(lr){2-3} \cmidrule(lr){4-5} \cmidrule(lr){6-7} \cmidrule(lr){8-9}
 & \textbf{Zero-shot} & \textbf{Few-shot} & \textbf{Zero-shot} & \textbf{Few-shot} & \textbf{Zero-shot} & \textbf{Few-shot} & \textbf{Zero-shot} & \textbf{Few-shot} \\
\midrule
Noise Ratio (Original)     & 57.39 & 56.66 & 57.70 & 55.78 & 60.61 & 61.49 & 44.94 & 44.42 \\
No Answer Ratio            & 0.00 & 0.00 & 0.001 & 0.0005 & 0.00 & 0.00 & 0.009 & 0.001 \\
Noise Ratio (After RA)     & 57.39 & 56.66 & 57.75 & 55.78 & 60.61 & 61.49 & 44.94 & 44.42 \\
\bottomrule
\end{tabular}
\caption{Label noise ratio of SemEval training set by four LLMs. "RA": random assignment.}
\label{tb:LLMs_noise_ratio}
\end{table*}

\section{LLM-generated Noise vs Synthetic Noise vs Real-world Noise}
\label{ap:noise_dist}
\begin{figure*}[h]
    \includegraphics[width=\textwidth]{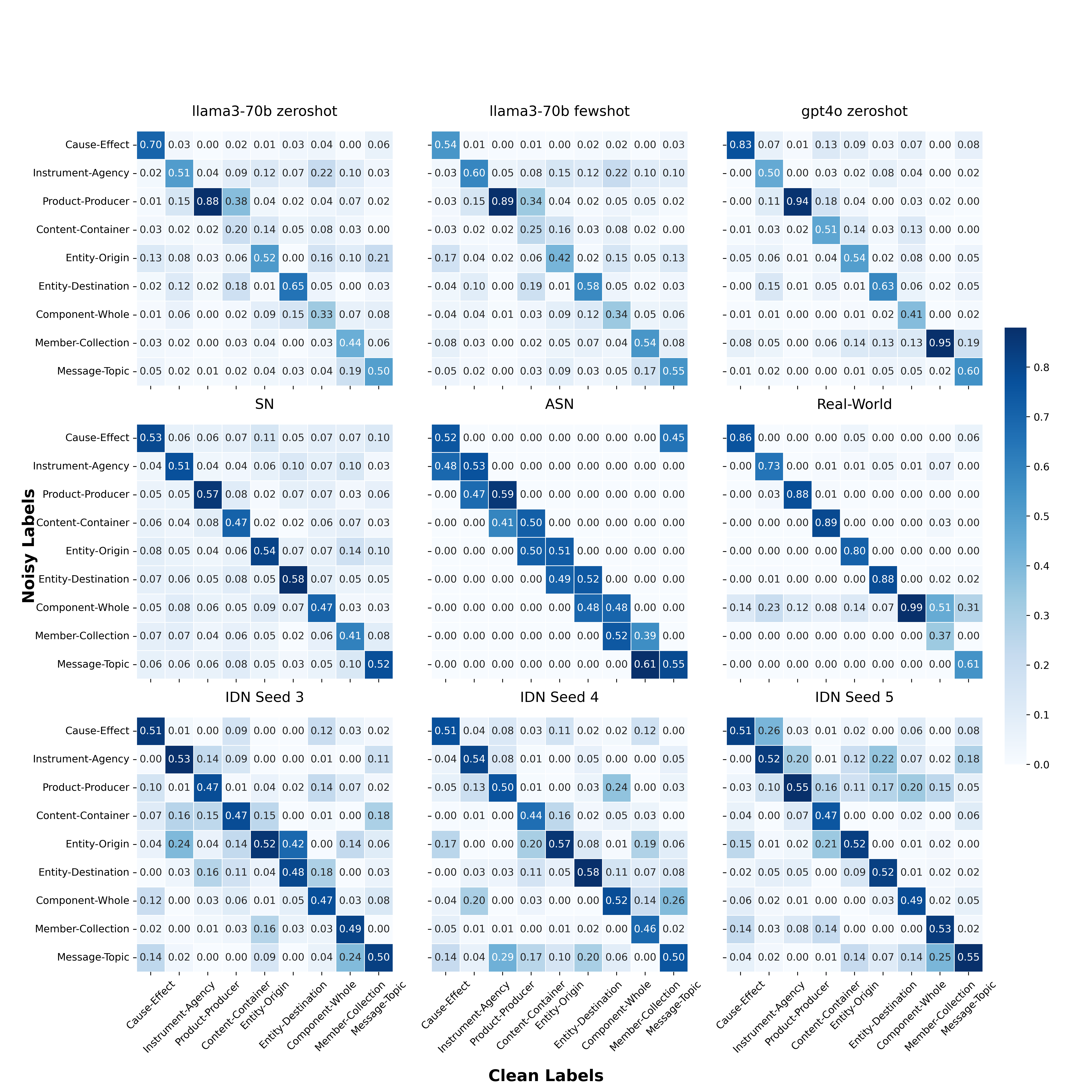}
    \caption{Confusion Matrix of LLM-generated label noise, synthetic noise, real-world noise on SemEval dataset. We include zeroshot and fewshot Llama-3-70b and zeroshot GPT4 for LLM-generated label. We use symmetric, asymmetric, and instance-dependent noise under three seeds for synthetic noise. Real-world noise is collected by 164 labeling functions written by subject matter expert.}
    \label{fig:noise_character_full}
\end{figure*}

We show the noise distribution comparison among LLMs, synthesis, and real-world in Figure \ref{fig:noise_character_full}.
